\documentclass[sigconf, nonacm]{acmart}

\newcommand\vldbdoi{XX.XX/XXX.XX}
\newcommand\vldbpages{XXX-XXX}
\newcommand\vldbvolume{14}
\newcommand\vldbissue{1}
\newcommand\vldbyear{2020}
\newcommand\vldbauthors{\authors}
\newcommand\vldbtitle{\shorttitle} 
\newcommand\vldbavailabilityurl{URL_TO_YOUR_ARTIFACTS}
\newcommand\vldbpagestyle{plain} 

\begin{document}
\title{AntBatchInfer: Elastic Batch Inference in the Kubernetes Cluster}

\author{Siyuan Li,
Youshao Xiao,
Fanzhuang Meng,
Lin Ju,
Lei Liang,
Lin Wang,
Jun Zhou
}
\affiliation{%
  \institution{Ant Group, Hangzhou, China}
  \city{}
  \country{}}
\email{{lisiyuan.li,youshao.xys, mengfanzhuang.mfz, julin.jl,leywar.liang,fred.wl, jun.zhoujun}@antgroup.com}

\begin{abstract}
Offline batch inference is a common task in the industry for deep learning applications, but it can be challenging to ensure stability and performance when dealing with large amounts of data and complicated inference pipelines. This paper demonstrated AntBatchInfer, an elastic batch inference framework, which is specially optimized for the non-dedicated cluster. AntBatchInfer addresses these challenges by providing multi-level fault-tolerant capabilities, enabling the stable execution of versatile and long-running inference tasks. It also improves inference efficiency by pipelining, intra-node, and inter-node scaling. It further optimizes the performance in complicated multiple-model batch inference scenarios. Through extensive experiments and real-world statistics, we demonstrate the superiority of our framework in terms of stability and efficiency. In the experiment, it outperforms the baseline by at least $2\times$ and  $6\times$ in the single-model or multiple-model batch inference. Also, it is widely used at Ant Group, with thousands of daily jobs from various scenarios, including DLRM, CV, and NLP, which proves its practicability in the industry.

\end{abstract}

\maketitle

\pagestyle{\vldbpagestyle}
\begingroup\small\noindent\raggedright\textbf{PVLDB Reference Format:}\\
\vldbauthors. \vldbtitle. PVLDB, \vldbvolume(\vldbissue): \vldbpages, \vldbyear.\\
\href{https://doi.org/\vldbdoi}{doi:\vldbdoi}
\endgroup
\begingroup
\renewcommand\thefootnote{}\footnote{\noindent
This work is licensed under the Creative Commons BY-NC-ND 4.0 International License. Visit \url{https://creativecommons.org/licenses/by-nc-nd/4.0/} to view a copy of this license. For any use beyond those covered by this license, obtain permission by emailing \href{mailto:info@vldb.org}{info@vldb.org}. Copyright is held by the owner/author(s). Publication rights licensed to the VLDB Endowment. \\
\raggedright Proceedings of the VLDB Endowment, Vol. \vldbvolume, No. \vldbissue\ %
ISSN 2150-8097. \\
\href{https://doi.org/\vldbdoi}{doi:\vldbdoi} \\
}\addtocounter{footnote}{-1}\endgroup

\ifdefempty{\vldbavailabilityurl}{}{
\vspace{.3cm}
\begingroup\small\noindent\raggedright\textbf{PVLDB Artifact Availability:}\\
The source code, data, and/or other artifacts have been made available at \url{\vldbavailabilityurl}.
\endgroup
}

\section{Introduction}

In the industry, the deployment of deep learning models can be categorized into two types: offline inference (batch inference) and online inference. In contrast to latency-sensitive online inference, the batch inference is less sensitive to latency but requires high throughput. This makes it ideal for massive business workloads that do not require immediate prediction results, which are also ubiquitous in the industry. For example, a use case is that batch inference enables full-graph inference of industrial-scale graph neural networks which may contain millions or even billions of nodes, to discover the potential social relationships~\cite{zhangagl}.

Unfortunately, most existing works and systems are devoted to online inference, while there have been few systematic works that consider batch inference in production, which is also crucial to industrial applications. One direct approach is to apply the online inference pipeline to batch jobs. However, offline inference has unique characteristics that distinguish it from online inference, such as massive latency-insensitive workloads and cost control~\cite{azure_batch_infer}. For example, samples to be processed could be up to terabytes in the industry. Another approach is to use batch processing systems such as MapReduce and Spark, which are able to process massive datasets with the guarantee of both efficiency and fault tolerance. However, they do not well suit large or complicated model inferences. For example, Deep Learning Recommendation Models usually need to store the large sparse parameters over several parameter servers~\cite{li2014scaling}. Moreover, the MapReduce-like batch processing systems are inflexible when it comes to executing the complicated multiple-model inference pipelines where the model complexity varies. Therefore, a solution is to train and serve these models in the container cluster, such as the Kubernetes cluster.

However, there are two primary problems posed to the existing batch inference systems in the K8S cluster: stability (fault tolerance) and efficiency. The conventional approach is to evenly distribute the total dataset to all the workers in containers and perform the model computations in a data-parallel fashion. Firstly, fault tolerance is vital in batch inference in the non-dedicated cluster (or shared cluster) at scale, where the scheduler may evict batch jobs to ensure the SLA of online jobs~\cite{bernstein2014containers}. While most batch inference systems provided by cloud vendors~\cite{aws_sagemaker,vertex_inference} provide pod-level fault tolerance and intra-node elasticity (elastically scaling out or in nodes) but do not consider application-level fault tolerance in the runtime, such as loading errors, or timeout errors. Secondly, these systems do not fully utilize computing resources, particularly in multiple-model inference scenarios or ensemble methods. A typical solution is to assign a model predictor process to a GPU device, which triggers resource waste when the model is too simple to utilize the GPU fully. Additionally, consider a multiple-model batch inference scenario, e.g., face recognition. Given the same input images, the customers require to perform the object detection stage, following the image classification stage to predict once. However, these two models are ensembled in the same predictor process in the current systems, such as Azure batch transform~\cite{azure_batch_infer}, and Google's Vertex~\cite{vertex_inference} while these two models have heterogeneous workloads.

Therefore, we demonstrate a k8s-based batch inference system that systematically addresses the stability and performance issues from the framework view. Firstly, a fine-grained fault-tolerant mechanism has been designed to ensure stability throughout the inference pipeline. Secondly, we
propose a pipelining to fully utilize the computing resources with intra-node and inter-node scaling for both single-model and multiple-model batch inference. Lastly, we demonstrate the simple user interface integrated with multiple backends and verify the superiority of our system in stability and efficiency.

\section{Problem Analysis}
 Consider a typical batch inference pipeline that consists of data ingestion, data wrangling, model inference, and persisting results. The data ingestion module reads samples from multiple data sources such as object storage and database systems. Then the data wrangling module preprocesses the samples, performing tasks like tokenization in NLP scenarios or data augmentation in CV scenarios, followed by model inference. Lastly, the prediction results are written back into the storage system for further usage in downstream applications. The distributed batch inference usually keeps a copy of the whole model parameters on each node and performs the batch inference given the pre-partitioned subdataset.
 
Let us further analyze the potential stability and efficiency problems throughout the inference pipeline. In terms of stability, there is a high risk of failures during the long-running job execution, leading to frequent failovers. This significantly damages the inference efficiency, particularly for batch jobs in non-dedicated clusters. We classify these failures into pod failures, application failures, and data failures. Firstly, we observe that the pod-level failures come from hardware failures, IO connection failures, and job preemption. Secondly, deep learning applications may encounter several potential problems when processing large datasets, including NAN values in samples, parsing errors, and hanging processes. These failures are ubiquitous throughout the whole procedure, but they are different from pod-level failures, which do not need heavy-weight pod-level failover. Lastly, the data fault tolerance should be carefully designed; otherwise, the data may be lost or duplicated during the pod failover, which harms the data integrity.

Also, the current system design presents challenges in achieving optimal performance for batch inference tasks. Firstly, the IO module such as data ingestion and writing back is heavy on data I/O, while the data wrangling and model inference are computation-intensive but have differences. The model inference is usually GPU-centric while the data wrangling is CPU-centric in most cases. Therefore, it is inefficient to ensemble these IO-intensive and GPU-centric or CPU-centric operations in the same module, otherwise, the pipeline is likely to bottleneck. Secondly, there are several models with different model complexity in the multiple-model batch inference scenarios, which is also inefficient to encapsulate them in the same module. Thirdly, batch inference usually runs on non-dedicated clusters where stragglers are common due to resource contention at peak periods. This leads to the long-tailed node problem where even data partition strategy results in the idleness of fast nodes when accomplishing their pre-assigned dataset but having to wait for the straggler nodes. However, the job completion time is decided by the slowest node. Lastly, idle computing resources in the non-dedicated cluster could be used to speed up batch jobs during the valleys period.

\begin{figure}
  \centering
  \includegraphics[width=0.9\linewidth]{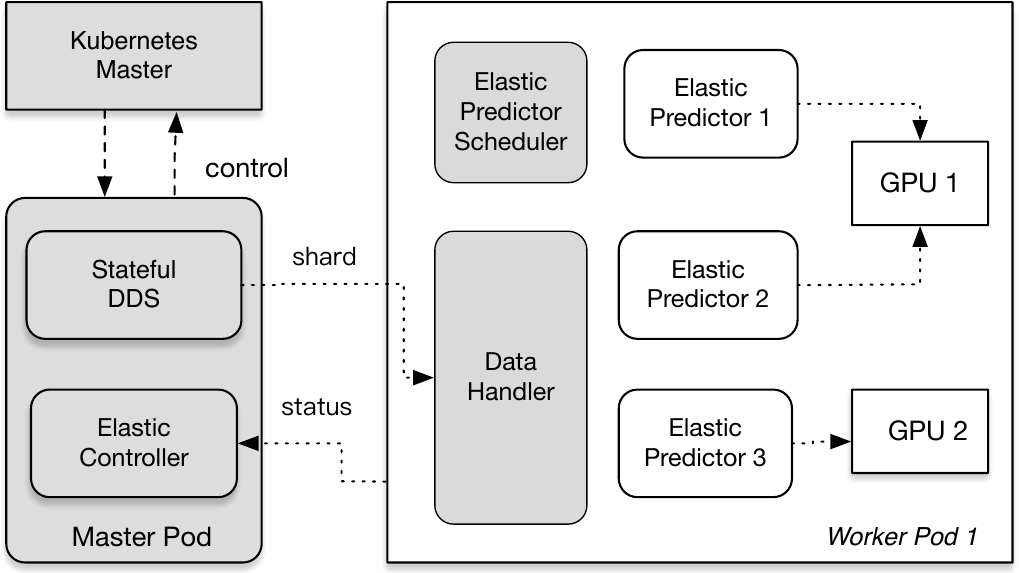}
  \caption{The System overview of the AntBacthInfer}
  \label{fig:arch}
\end{figure}

\section{Our Framework}

\subsection{Framework Architecture}
To ensure the stability and efficiency of batch inference, we propose the AntBatchInfer framework. As shown in Figure \ref{fig:arch}, this framework comprises four modules: Stateful Data Sharding Service (Stateful DDS), Data Handler, Elastic Controller, and Elastic Predictor Scheduler. The system is designed with a master-worker architecture, where Stateful DDS and Elastic Controller are located on a separate master node, while the remaining modules are situated on each worker node.

\textbf{Stateful Data Sharding Service (Stateful DDS)} elastically distributes the data samples to each worker with unbalanced computation capacity and manages the lifecycle of data samples at the shard level. On one hand, the stateful DDS maintains a global message queue where the entire dataset is partitioned at the shard level, and insert all data shards into the queue for workers to consume. Each data shard only contains metadata that records the index of samples in the storage system, and a shard may contain multiple batches. This approach helps to rebalance the workloads between fast and slow nodes, which solves the long-tailed node problems compared with the even data partition strategy. On the other hand, the DDS service also hosts the state information to trace the completion status of each shard, which helps the data fault tolerance in the inter-node scaling and failovers. These states are classified into three categories: "TODO", "DOING", and "DONE". All of the state transitions are conducted by the DDS service.

\textbf{Data Handler} is responsible for both I/O module and CPU-intensive data wrangling. It also collaborates with the stateful DDS for data loading and synchronizing the state of data shards. Specifically, the Data Handler in each node fetches the actual samples from multiple data sources according to the metadata in the shards assigned by the stateful DDS. It then preprocesses the data samples and put the results in a message queue for further model inference. Additionally, it is further optimized for small-file scenarios by merging small files and near caching in advance before the inference. Lastly, it reports the completion status of the data shard after committing the prediction results to the storage system.  

\textbf{Elactic Controller} plays a vital role in resource management at the node level throughout the batch inference job, including pod-level fault tolerance. It manages the lifecycle of all pods via communicating with the Kubernetes Master, including requesting computing resources, starting off the worker pod, and restarting the terminated worker pod if necessary. Additionally, the Elastic Controller allows for elastic scaling out or in the computing nodes by periodically querying the Kubernetes Master for computing resources on demand. This helps to speed up the batch inference job during the valley hours. In case of any retryable failures, such as hardware failures and job preemptions, the Elastic Controller can migrate the batch inference from crashed nodes to the new nodes with the assistance of the stateful DDS to ensure the data is neither lost nor duplicated.

\textbf{Elactic Predictor Scheduler} elastically scales out intra-node predictors which 
encapsulates the computation-intensive model computation logic. This elastic predictor is designed for three purposes. Firstly, it controls the process-level concurrency and adaptively scales out the data loading, predictor, and writing processes or threads to improve the utilization of computing resources. Secondly, it manages the lifecycle of these processes for fine-grained fault tolerance. These include the reboot of the hanging processes and reboot processes for unpredicted memory leaks in user code. Lastly, we enable different levels of parallelism among model predictors in the multiple-model batch inference and utilize the queue for communication.  


\begin{figure}
  \centering
  \includegraphics[width=0.9\linewidth]{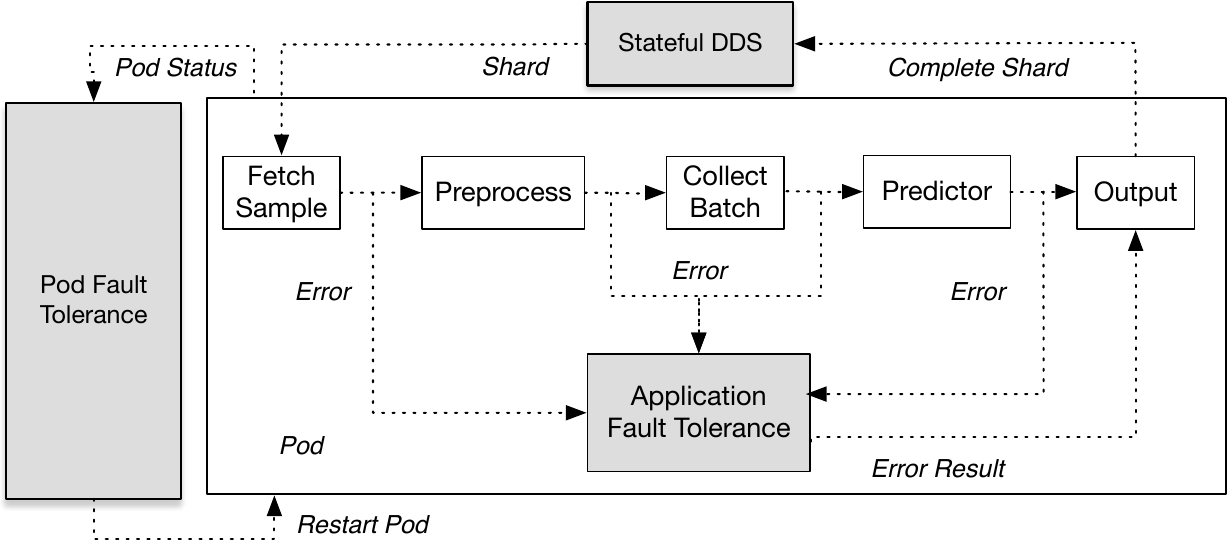}
  \caption{The multi-level fault tolerance.}
  \label{fig:fault_tolerance}
\end{figure}

\subsection{Optimization for the Stability}
This subsection explains the multi-level fault tolerance capability in AntBatchInfer as shown in Figure \ref{fig:fault_tolerance}. We classify our fault tolerance ability into three levels: pod fault tolerance, application fault tolerance, and data fault tolerance.

\subsubsection{Pod Fault Tolerance.}
The Elastic Controller periodically listens to the pod events among all nodes via the Kubernetes Master and classifies these node crashes into two retryable and unretryable errors. Typical retryable errors are network errors, hardware failures, and task eviction. Unretryable errors are configuration errors or programming errors from users. It starts off a new pod and launches the local batch inference task with the new data shard pulled from the stateful DDS for the retryable errors or scaling up events and terminates the pod for scaling down events.

\subsubsection{Application Fault Tolerance.} The Elastic Predictor Scheduler locally monitors the status of the processes during the batch inference. As depicted in Figure \ref{fig:fault_tolerance}, we first catch the errors encountered throughout the pipeline, including data fetching errors, parsing errors, or inference errors, and skip those tolerable errors. We associate them with the corresponding samples, package them in batches and write them all into the storage system. These tolerable error information in the output results helps the users with error analysis. Secondly, we restart the processes when encountering unforeseen errors with the timeout-retry mechanism. These issues include hanging processes and memory leaks caused by user code from various user cases.

\subsubsection{Data Fault Tolerance.} The Data Handler in the new worker first fetches "TODO" shards from the DDS service and reads samples from the data source. The shard is marked as "DOING" when batch inference begins. After that, the Predictor performs model computation based on the pulled data shard. The Data Handler reports the shard state after the prediction results have been committed into the storage system, and DDS marks these shards as "DONE" when it receives notifications from the Data Handler. When the Elastic Controller discovers any node crash caused by failures or scaling events, the assigned "DOING" shard to the worker will be marked as "TODO" by the DDS service, and DDS inserts the shard back into the end of the data queue. In this way, we guarantee data integrity in failovers or elasticity.

\subsection{Optimization for the Efficiency}

\subsubsection{Reducing the overall JCT}
Stateful DDS reduces the overall job completion time (JCT) and saves up computing resources by elastically allocating data samples to each worker based on the real-time throughput. This naturally achieves load-balancing among workers. It reduces the overall job completion time which is decided by the slowest machines compared with the even data partition strategy in the long-tailed node problem. Additionally, the elastic controller scales out more worker nodes to improve the training efficiency when the cluster is idle.

\subsubsection{Speedup Single-model Batch Inference} 
AntBatchInfer improves the single-model batch inference by decoupling it into three stages: data loading, prediction, and writing. These stages are encapsulated in separate threads or processes and the intra-node scheduler auto-scales these stages in different levels of concurrency based on a heuristic. It overlaps the execution of these stages (defined by users) in the timeline and these stages coordinate through a lock-free queue. Specifically, the scheduler increases the number of data-loading processes or threads when the message queue for model inference is almost empty and increases the number of model predictors when the message queue is full and CPU/GPU is underutilized. The writing thread is increased when the writing queue is full due to the long sequence output. This approach interleaves these IO-intensive and CPU or GPU-intensive workloads in the batch inference pipeline to maximize the throughput.

\subsubsection{Speedup Multiple-model Batch Inference Pipeline.} To enhance the efficiency of the Multiple-model Batch Inference, we propose encapsulating these models into several predictors, which comprise a DAG. Each predictor is a separate process consisting of a single model inference logic defined by users and could be elastically assigned the number of GPUs according to the model complexity. Additionally, the subsequent predictor immediately performs batch inference upon reaching the target batch size in our pipeline. We ensemble the results before outputting the results via the queue in shared memory. This avoids frequent initializations of the CUDA runtime when the batch size of model input varies. For instance, the object detection model may output a different number of semantic objects, which will be used in the subsequent classification model. 

\begin{figure}
  \centering
  \includegraphics[width=\linewidth]{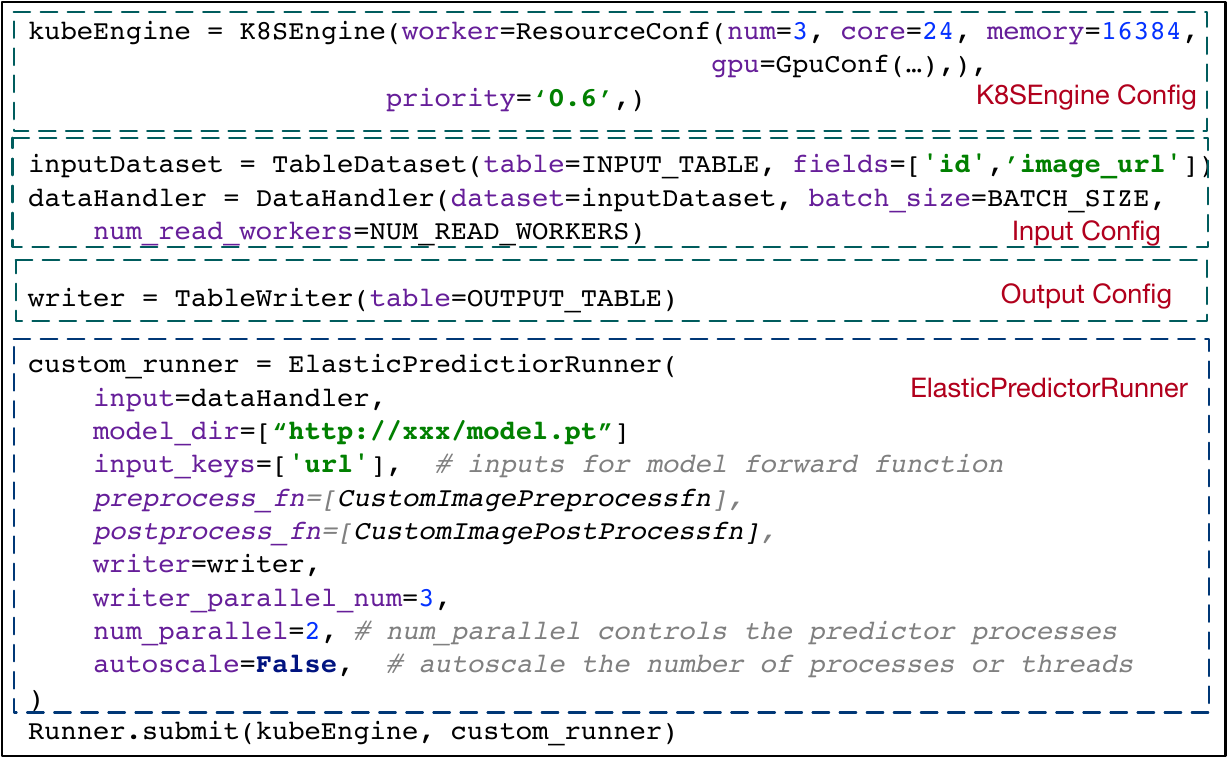}
  \caption{The user interface of AntBatchInfer.}
  \label{fig:interface}
\end{figure}

\section{Demonstration}
In our demonstration, we show the simple user interface of AntBatchInfer and demonstrate a use case using batch inference on an image classification task over AntBatchInfer as shown in figure 3. These configurations could be easily further applied to other batch inference tasks. 1) The {\sffamily EngineConfig} specifies the hardware resources. The users could specify the priority parameter to enable the inter-node elasticity, where 0.6 represents that 60\% of the computing resource is on-demand and others are spot resources. 2) The {\sffamily DataHandler} object provides the configuration of the data source, including the num\_workers to control the concurrency. It overrides the DataLoader object in Pytorch and Dataset object in Tensorflow. 3) The {\sffamily WriterConfig} specifies the target output storage system and the number of writing threads. 4) The {\sffamily ElasticPredictionRunner Config} specifies the model file and the number of predictors. Users could specify a list of preprocessing functions, postprocessing functions, and models for Single Model or Multiple-models Batch Inference. Note that users could manually specify the number of processes or turn on the intra-node autoscale feature. Additionally, we have a web-based GUI that allows users with little programming experience to effectively use AntBatchInfer. More details will be demonstrated in the further demo videos.

\begin{figure}
\begin{minipage}[t]{0.44\linewidth}
\includegraphics[width=\linewidth]{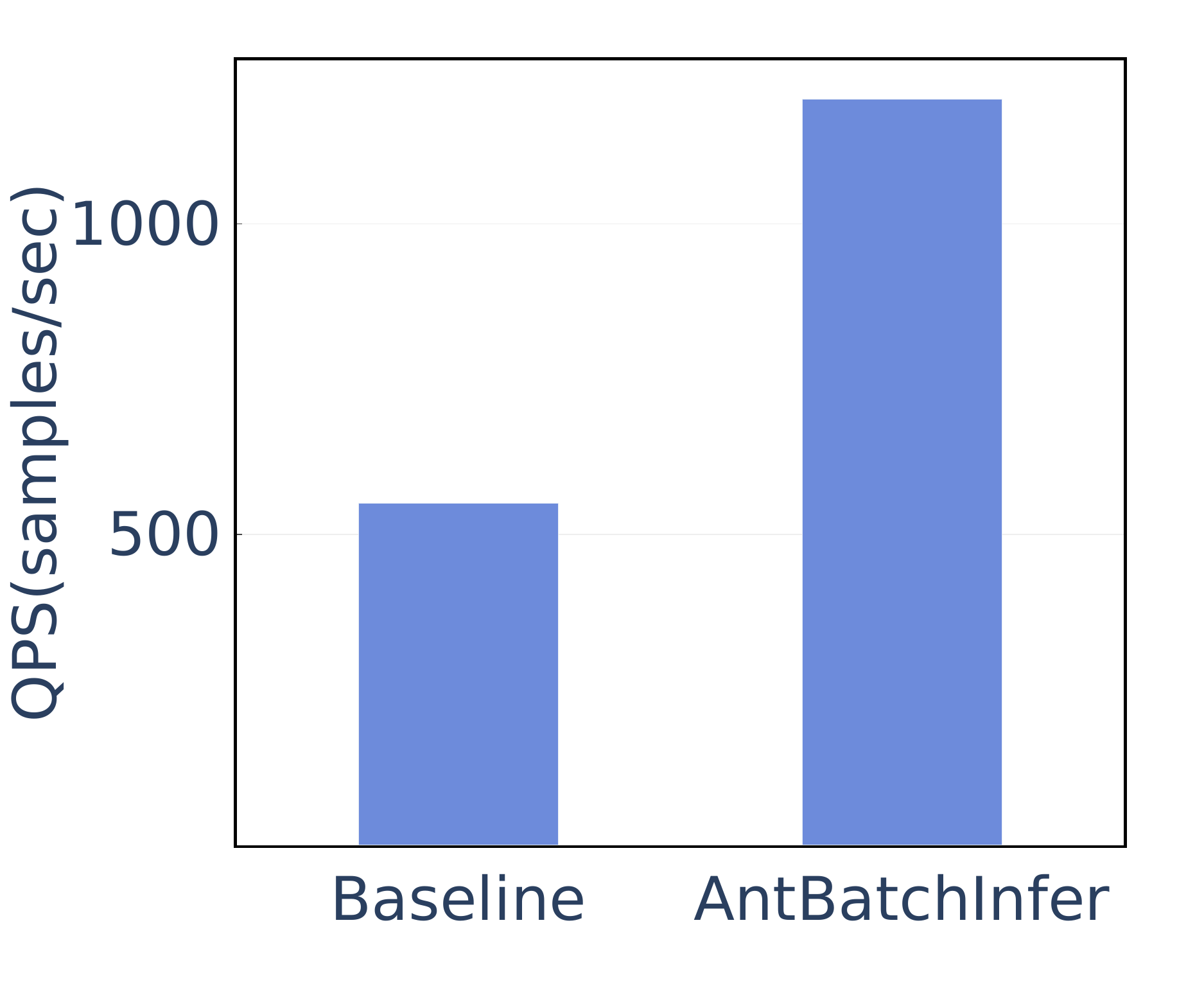} %
\caption{The throughput between baseline and AntBatchInfer in Single-model Batch Inference.}
\label{fig:singlemodel}
\end{minipage}
\hfill
\begin{minipage}[t]{0.44\linewidth}
\includegraphics[width=\linewidth]{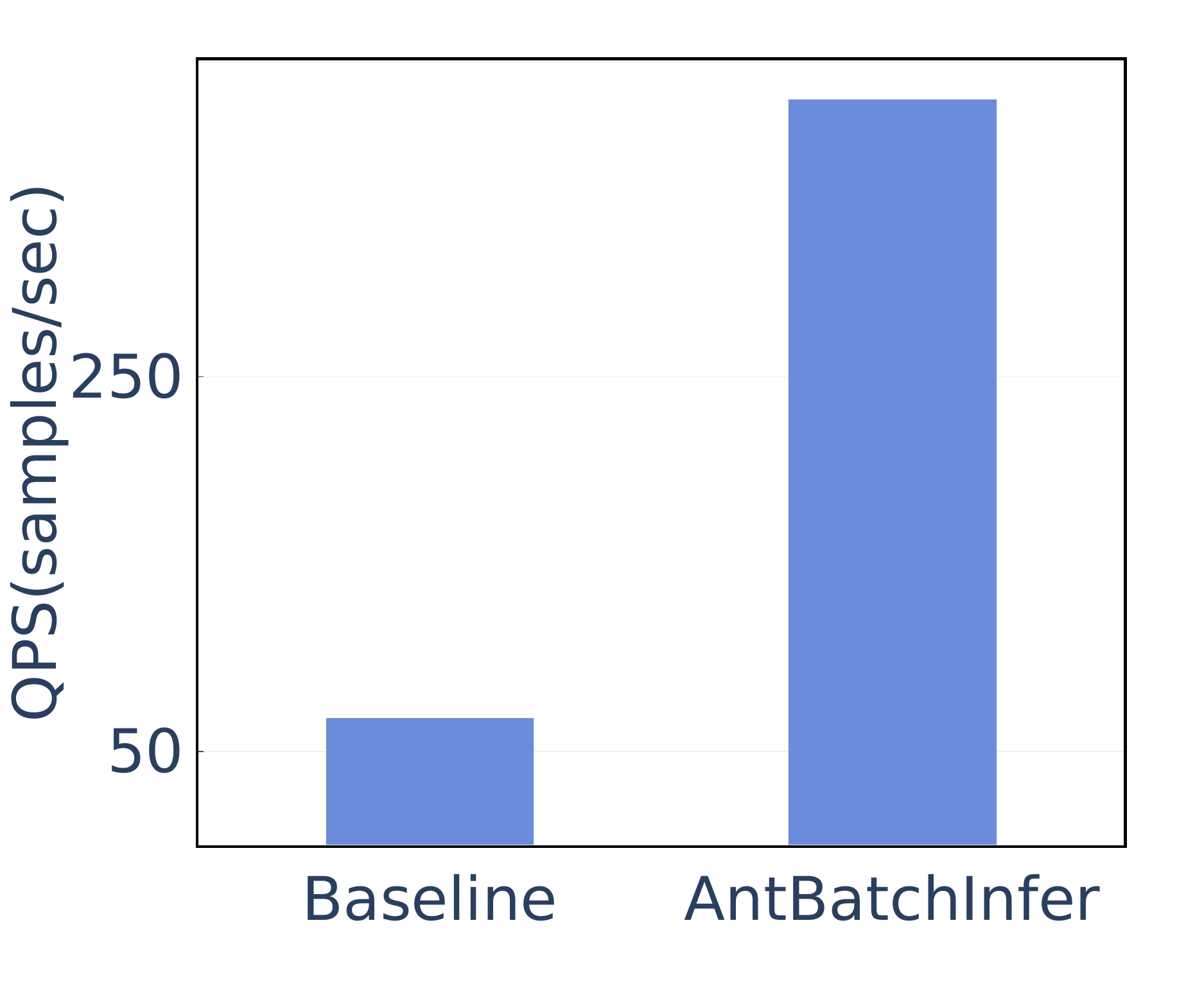} %
\caption{The throughput between baseline and AntBatchInfer in Multiple-model Batch Inference.}
\label{fig:multiplemodel}
\end{minipage}%
\end{figure}

\begin{figure}
\begin{minipage}[t]{0.46\linewidth}
\includegraphics[width=\linewidth]{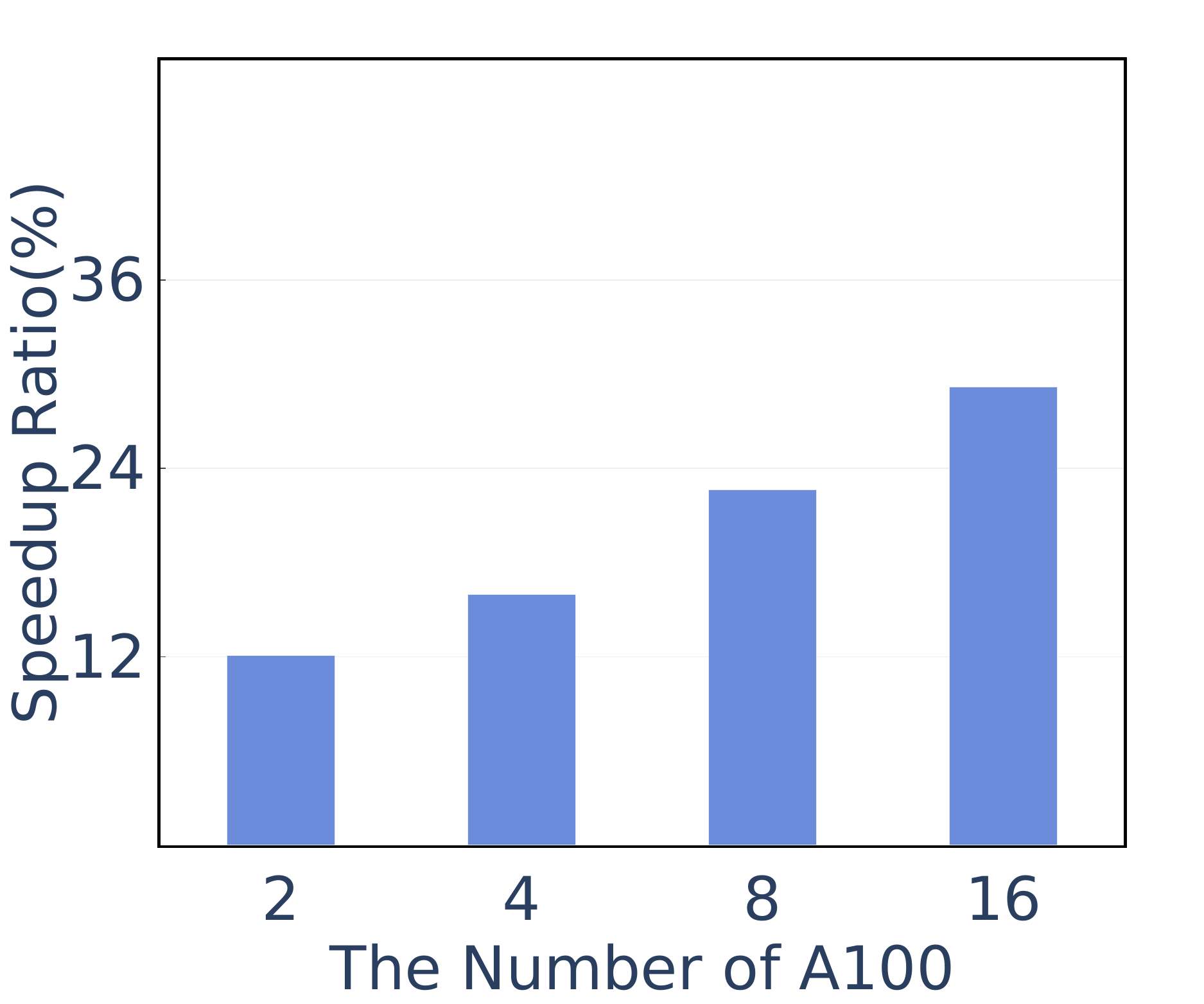} %
\caption{The speedup ratio between DDS and native data partition strategy.}
\label{fig:dds}
\end{minipage}
\hfill
\begin{minipage}[t]{0.46\linewidth}
\includegraphics[width=\linewidth]{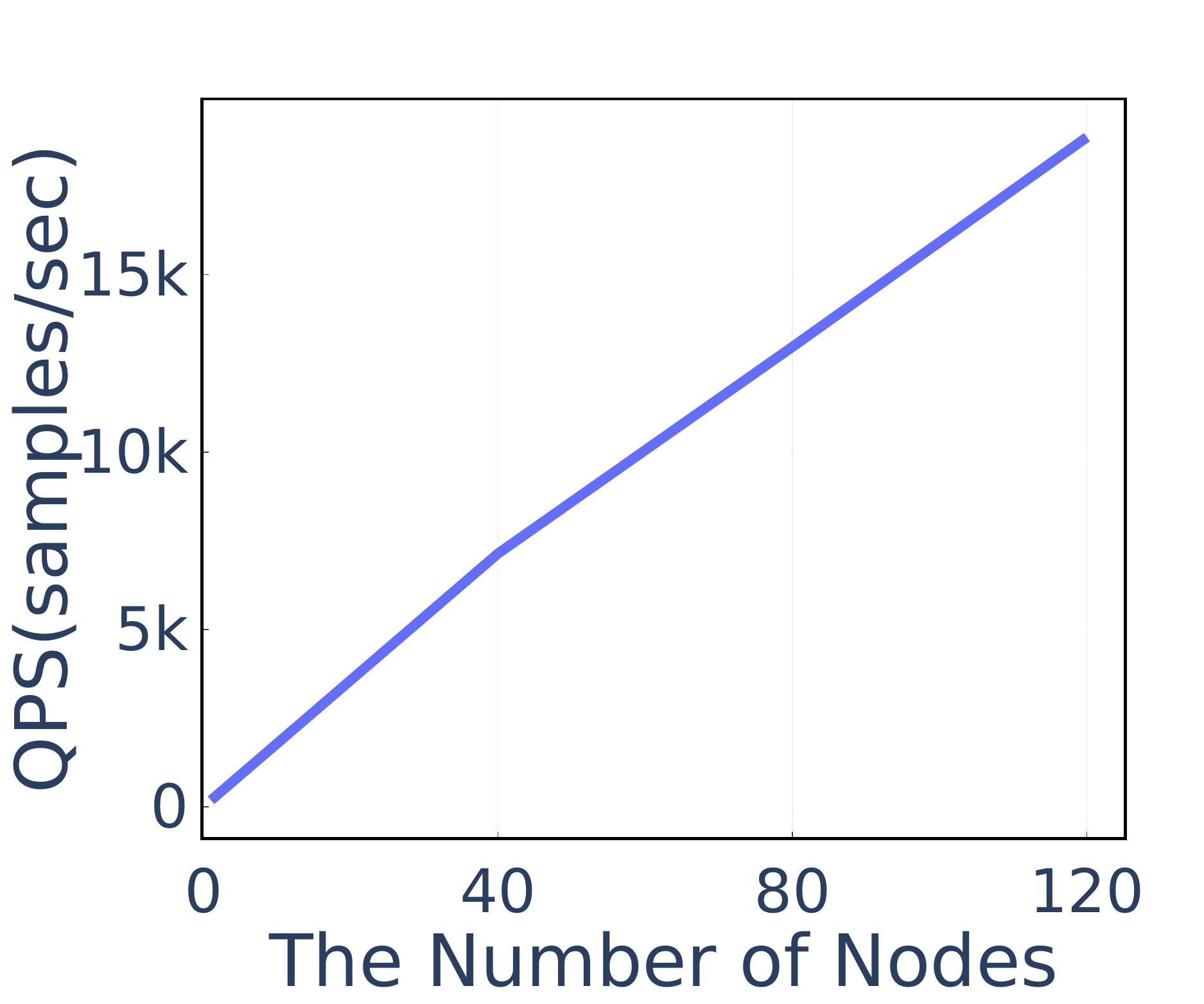} %
\caption{The throughput of AntBatchInfer when scaling out.}
\label{fig:scalability}
\end{minipage}%
\end{figure}

\section{Experiments}
In this section, we show the efficiency of AntBatchInfer, with further demonstrations of its multi-level fault tolerance and elasticity available in demo videos that use TensorFlow, PyTorch, and ONNX backend. Firstly, we evaluate AntBatchInfer's performance in a single-model batch inference job for a graph neural network, TGAT~\cite{xu2020inductive}, with half a billion nodes and 6 billion edges. This job performs batch inference for 260 million samples over a non-dedicated CPU cluster every day. Figure \ref{fig:singlemodel} shows that AntBatchInfer is at least two times faster than the baseline in terms of QPS, which is 550 and 1200, respectively. The baseline disables the pipeline workflow and intra-node autoscale feature. Secondly, we perform batch inference in a multiple-model scenario over Nvidia A100s, where the first stage is object detection using a variant of SCRFD model~\cite{guo2021sample}, and the second stage is image classification based on ResNet~\cite{he2016deep}. Figure \ref{fig:multiplemodel} displays that AntBatchInfer achieves nearly six times faster QPS than the baseline, which is 68 and 398, respectively. The baseline packages two models in one stage sequentially. Thirdly, we further compare the job completion time between the even data strategy and the DDS-based method in the multiple-model scenario. Figure \ref{fig:dds} shows that the DDS-based method achieves 12\% to 30\% speedup against the baseline even over A100s. This is because the various complexity of data and model complexity make the difference. The gap is much larger in the non-dedicated cluster according to our experience. Lastly, Figure \ref{fig:scalability} shows that AntBatchInfer scales linearly when adding up to 120 CPU nodes, where each node owns 20 cores. This verifies that the synchronization cost between stateful DDS and the worker nodes is negligible.

\bibliographystyle{ACM-Reference-Format}
\bibliography{sample}


\begin{thebibliography}{9}


\ifx \showCODEN    \undefined \def \showCODEN     #1{\unskip}     \fi
\ifx \showDOI      \undefined \def \showDOI       #1{#1}\fi
\ifx \showISBNx    \undefined \def \showISBNx     #1{\unskip}     \fi
\ifx \showISBNxiii \undefined \def \showISBNxiii  #1{\unskip}     \fi
\ifx \showISSN     \undefined \def \showISSN      #1{\unskip}     \fi
\ifx \showLCCN     \undefined \def \showLCCN      #1{\unskip}     \fi
\ifx \shownote     \undefined \def \shownote      #1{#1}          \fi
\ifx \showarticletitle \undefined \def \showarticletitle #1{#1}   \fi
\ifx \showURL      \undefined \def \showURL       {\relax}        \fi
\providecommand\bibfield[2]{#2}
\providecommand\bibinfo[2]{#2}
\providecommand\natexlab[1]{#1}
\providecommand\showeprint[2][]{arXiv:#2}

\bibitem[\protect\citeauthoryear{Bernstein}{Bernstein}{2014}]%
        {bernstein2014containers}
\bibfield{author}{\bibinfo{person}{David Bernstein}.}
  \bibinfo{year}{2014}\natexlab{}.
\newblock \showarticletitle{Containers and cloud: From lxc to docker to
  kubernetes}.
\newblock \bibinfo{journal}{\emph{IEEE Cloud Computing}} \bibinfo{volume}{1},
  \bibinfo{number}{3} (\bibinfo{year}{2014}), \bibinfo{pages}{81--84}.
\newblock


\bibitem[\protect\citeauthoryear{Google~Cloud}{Google~Cloud}{2020}]%
        {vertex_inference}
\bibfield{author}{\bibinfo{person}{Inc Google~Cloud}.}
  \bibinfo{year}{2020}\natexlab{}.
\newblock \bibinfo{title}{Vertex AI}.
\newblock
\newblock
\urldef\tempurl%
\url{https://cloud.google.com/vertex-ai/docs/predictions/get-predictions}
\showURL{%
\tempurl}


\bibitem[\protect\citeauthoryear{Guo, Deng, Lattas, and Zafeiriou}{Guo
  et~al\mbox{.}}{2021}]%
        {guo2021sample}
\bibfield{author}{\bibinfo{person}{Jia Guo}, \bibinfo{person}{Jiankang Deng},
  \bibinfo{person}{Alexandros Lattas}, {and} \bibinfo{person}{Stefanos
  Zafeiriou}.} \bibinfo{year}{2021}\natexlab{}.
\newblock \bibinfo{title}{Sample and Computation Redistribution for Efficient
  Face Detection}.
\newblock
\newblock
\showeprint[arxiv]{2105.04714}~[cs.CV]


\bibitem[\protect\citeauthoryear{He, Zhang, Ren, and Sun}{He
  et~al\mbox{.}}{2016}]%
        {he2016deep}
\bibfield{author}{\bibinfo{person}{Kaiming He}, \bibinfo{person}{Xiangyu
  Zhang}, \bibinfo{person}{Shaoqing Ren}, {and} \bibinfo{person}{Jian Sun}.}
  \bibinfo{year}{2016}\natexlab{}.
\newblock \showarticletitle{Deep residual learning for image recognition}. In
  \bibinfo{booktitle}{\emph{Proceedings of the IEEE conference on computer
  vision and pattern recognition}}. \bibinfo{pages}{770--778}.
\newblock


\bibitem[\protect\citeauthoryear{Hudgeon and Nichol}{Hudgeon and
  Nichol}{2020}]%
        {aws_sagemaker}
\bibfield{author}{\bibinfo{person}{Doug Hudgeon} {and} \bibinfo{person}{Richard
  Nichol}.} \bibinfo{year}{2020}\natexlab{}.
\newblock \bibinfo{title}{Machine Learning for Business: Using amazon sagemaker
  and Jupyter}.
\newblock
\newblock
\urldef\tempurl%
\url{https://docs.aws.amazon.com/sagemaker/latest/dg/deploy-model.html}
\showURL{%
\tempurl}


\bibitem[\protect\citeauthoryear{Li et~al\mbox{.}}{Li et~al\mbox{.}}{2014}]%
        {li2014scaling}
\bibfield{author}{\bibinfo{person}{Mu Li} {et~al\mbox{.}}}
  \bibinfo{year}{2014}\natexlab{}.
\newblock \showarticletitle{Scaling distributed machine learning with the
  parameter server}. In \bibinfo{booktitle}{\emph{OSDI 14}}.
  \bibinfo{pages}{583--598}.
\newblock


\bibitem[\protect\citeauthoryear{Microsoft}{Microsoft}{2020}]%
        {azure_batch_infer}
\bibfield{author}{\bibinfo{person}{Inc Microsoft}.}
  \bibinfo{year}{2020}\natexlab{}.
\newblock \bibinfo{title}{Batch inference in azure machine learning}.
\newblock
\newblock
\urldef\tempurl%
\url{https://techcommunity.microsoft.com/t5/ai-machine-learning-blog/batch-inference-in-azure-machine-learning/ba-p/1417010}
\showURL{%
\tempurl}


\bibitem[\protect\citeauthoryear{Xu, Ruan, Korpeoglu, Kumar, and Achan}{Xu
  et~al\mbox{.}}{2020}]%
        {xu2020inductive}
\bibfield{author}{\bibinfo{person}{Da Xu}, \bibinfo{person}{Chuanwei Ruan},
  \bibinfo{person}{Evren Korpeoglu}, \bibinfo{person}{Sushant Kumar}, {and}
  \bibinfo{person}{Kannan Achan}.} \bibinfo{year}{2020}\natexlab{}.
\newblock \showarticletitle{Inductive representation learning on temporal
  graphs}.
\newblock \bibinfo{journal}{\emph{arXiv preprint arXiv:2002.07962}}
  (\bibinfo{year}{2020}).
\newblock


\bibitem[\protect\citeauthoryear{Zhang, Huang, Liu, Zhou, Hu, Song, Ge, Wang,
  Zhang, and Qi}{Zhang et~al\mbox{.}}{2020}]%
        {zhangagl}
\bibfield{author}{\bibinfo{person}{Dalong Zhang}, \bibinfo{person}{Xin Huang},
  \bibinfo{person}{Ziqi Liu}, \bibinfo{person}{Jun Zhou},
  \bibinfo{person}{Zhiyang Hu}, \bibinfo{person}{Xianzheng Song},
  \bibinfo{person}{Zhibang Ge}, \bibinfo{person}{Lin Wang},
  \bibinfo{person}{Zhiqiang Zhang}, {and} \bibinfo{person}{Yuan Qi}.}
  \bibinfo{year}{2020}\natexlab{}.
\newblock \showarticletitle{AGL: A Scalable System for Industrial-purpose Graph
  Machine Learning}.
\newblock \bibinfo{journal}{\emph{Proceedings of the VLDB Endowment}}
  \bibinfo{volume}{13}, \bibinfo{number}{12} (\bibinfo{year}{2020}).
\newblock


\end{thebibliography}

\end{document}